\pdfoutput=1

\documentclass[11pt]{article}

\usepackage[final]{acl}
\usepackage[tikz]{bclogo}
\usepackage{lipsum}
\usepackage{times}
\usepackage{latexsym}

\usepackage[T1]{fontenc} 

\usepackage[utf8]{inputenc}

\usepackage{microtype}

\usepackage{booktabs}
\usepackage{graphicx} 
\usepackage{graphics}
\usepackage{amssymb}
\usepackage{amsmath}
\usepackage{pifont}
\usepackage{makecell}
\usepackage{resizegather}
\usepackage{etoolbox}
\usepackage{booktabs,makecell,tabularx} 
\usepackage{enumitem}
\usepackage{mathtools}
\usepackage{cleveref}
\usepackage{multirow}
\usepackage{booktabs,array}
\usepackage{amsmath, bm}
\usepackage{color, colortbl}
\usepackage{xcolor}
\usepackage{url}
\usepackage{hyperref}
\usepackage{arydshln}
\usepackage{booktabs}
\usepackage{multirow}
\usepackage{fdsymbol}
\usepackage{tcolorbox}

\usepackage[ruled,vlined]{algorithm2e}         
\newcolumntype{P}[1]{>{\centering\arraybackslash}p{#1}}

\let\oldnl\nl
\definecolor{Gray}{gray}{0.9}
\definecolor{LightCyan}{rgb}{0.88,1,1}
\newcommand{\nonl}{\renewcommand{\nl}{\let\nl\oldnl}}

\newcolumntype{H}{>{\setbox0=\hbox\bgroup}c<{\egroup}@{}}

\newcommand*\inlineimage[1]{\raisebox{-0.15\baselineskip}{$\,$\includegraphics[height=0.9\baselineskip]{#1}$\,\,$}}
\newcommand{\blender}{\inlineimage{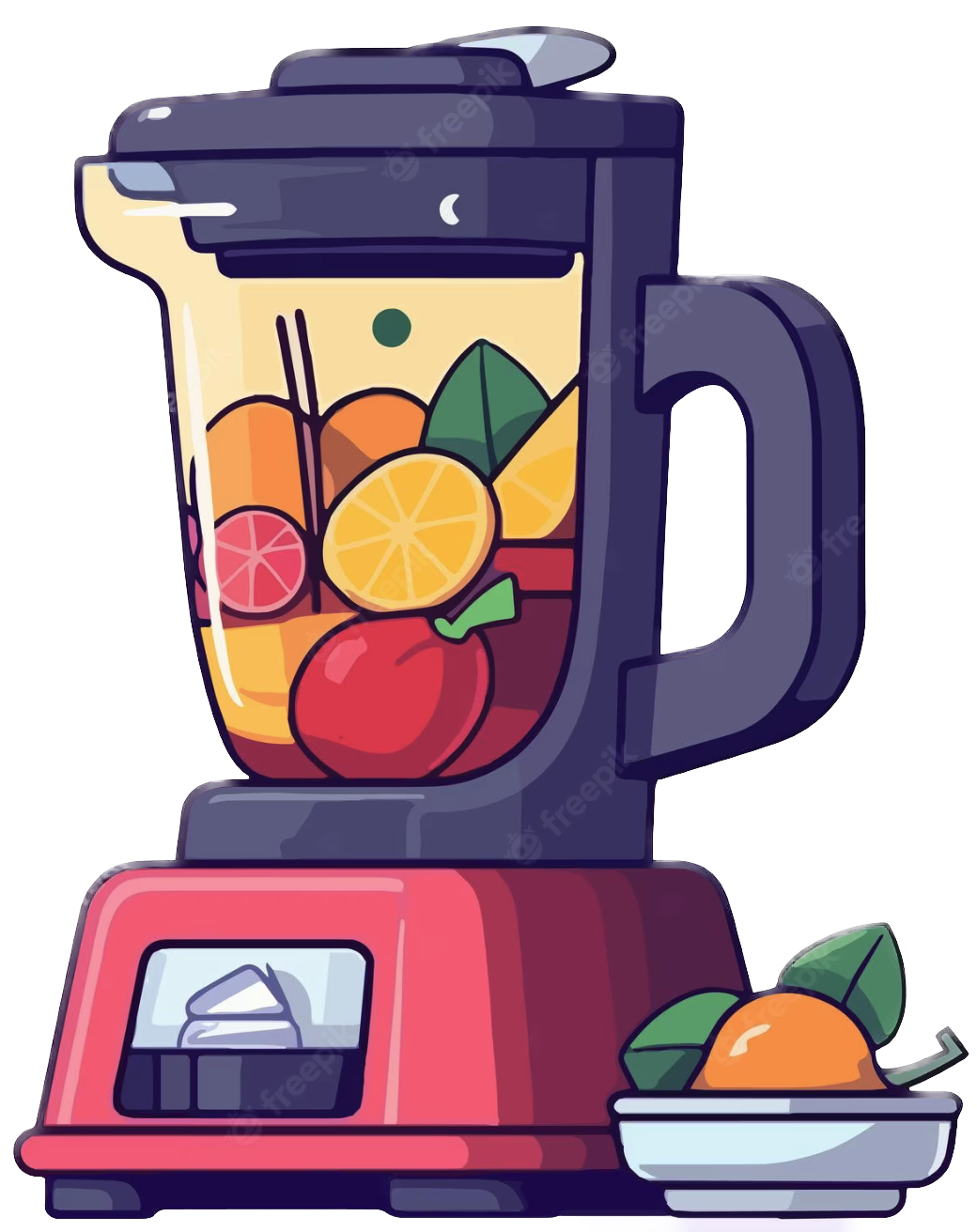}}

\newcommand*\inlinelargeimage[1]{\raisebox{-0.15\baselineskip}{$\,$\includegraphics[height=1.2\baselineskip]{#1}$\,\,$}}
\newcommand{\largeblender}{\inlinelargeimage{figures/blender2.png}}

\newcommand{\mixdata}{\inlineimage{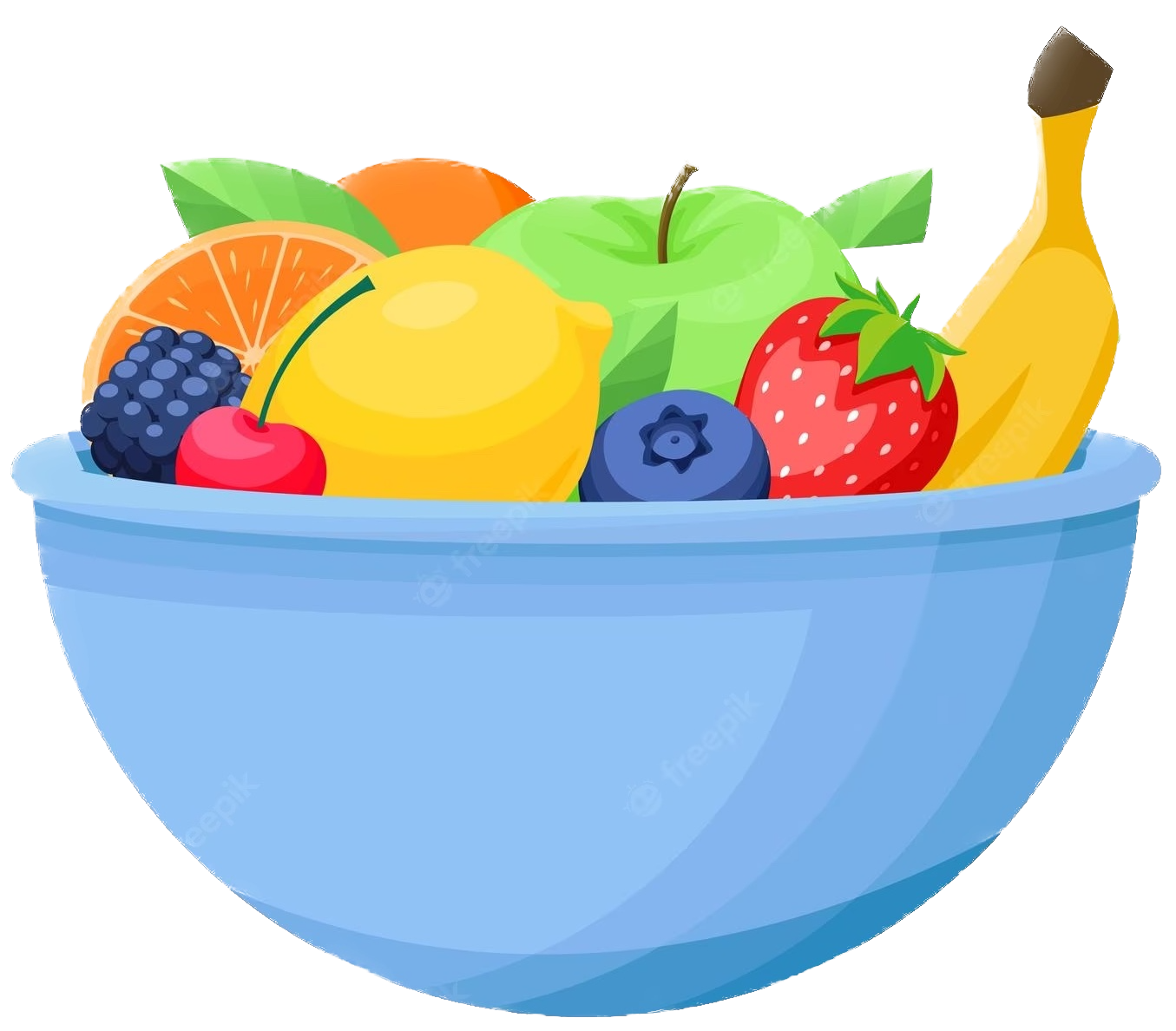}}

\newcommand{\methodname}[1]{\texttt{LLM-\textsc{Blender}}}{}
\newcommand{\dataname}[1]{\texttt{MixInstruct}}{}

\newcommand{\ranker}[1]{\textsc{PairRanker}}{}
\newcommand{\fuser}[1]{\textsc{GenFuser}}{}

\title{
\largeblender
\methodname{}: Ensembling Large Language Models \\
with Pairwise Ranking and Generative Fusion
}

\newcommand{\aspace}{\hspace{1em}}

\newcommand{\usc}{$^{\clubsuit}$}
\newcommand{\aiTwo}{$^{\spadesuit}$}
\newcommand{\zju}{$^{\heartsuit}$}

\author{
Dongfu Jiang\zju \aspace Xiang Ren\usc\aiTwo \aspace Bill Yuchen Lin\aiTwo 
\\
{\small{\texttt{dongfu@zju.edu.cn, xiangren@usc.edu, yuchenl@allenai.org} }} \\ 
\aiTwo Allen Institute for Artificial Intelligence \\ 
\usc University of Southern California \aspace
\zju Zhejiang University \quad  \\
}

\begin{document}
\maketitle

\begin{abstract}

We present \methodname{}, an ensembling framework designed to attain consistently superior performance by leveraging the diverse strengths of multiple open-source large language models (LLMs). Our framework consists of two modules: \textsc{PairRanker} and \textsc{GenFuser}, addressing the observation that optimal LLMs for different examples can significantly vary.
\textsc{PairRanker} employs a specialized pairwise comparison method to distinguish subtle differences between candidate outputs. It jointly encodes the input text and a pair of candidates, using cross-attention encoders to determine the superior one. Our results demonstrate that \ranker{} exhibits the highest correlation with ChatGPT-based ranking. 
Then, \textsc{GenFuser} aims to merge the top-ranked candidates, generating an improved output by capitalizing on their strengths and mitigating their weaknesses. To facilitate large-scale evaluation, we introduce a benchmark dataset, {\dataname{}}, which is a mixture of multiple instruction datasets featuring oracle pairwise comparisons.
Our \methodname{} significantly outperform individual LLMs and baseline  methods across various metrics, establishing a substantial performance gap.
\footnote{\url{https://yuchenlin.xyz/LLM-Blender}}~\footnote{The experiments on summarization, translation, and constrained generation tasks in the prior version have been moved to the appendix. Instead, we mainly present our work in the context of instruction-following data and LLMs in this version.}
\end{abstract}

\section{Introduction}
\label{sec:intro}
Large language models (LLMs) have shown impressive performance in diverse tasks, primarily due to their capacity to follow instructions and access extensive, high-quality data, showing a promising future for artificial general intelligence~\cite{Bubeck2023SparksOA}. 
However, prominent LLMs such as GPT-4 and PaLM~\cite{Chowdhery2022PaLMSL} are closed-source, restricting insights into their architectures and training data. Open-source LLMs like Pythia~\cite{Biderman2023PythiaAS}, LLaMA~\cite{Touvron2023LLaMAOA}, and Flan-T5~\cite{Chung2022ScalingIL} offer a chance to fine-tune these models on custom instruction datasets, enabling the development of smaller yet efficient LLMs, such as Alpaca, Vicuna~\cite{vicuna2023}, OpenAssistant~\cite{oasst}, and MPT~\cite{MosaicML2023Introducing}.

\begin{figure}
    \centering
    \includegraphics[width=0.95\linewidth]{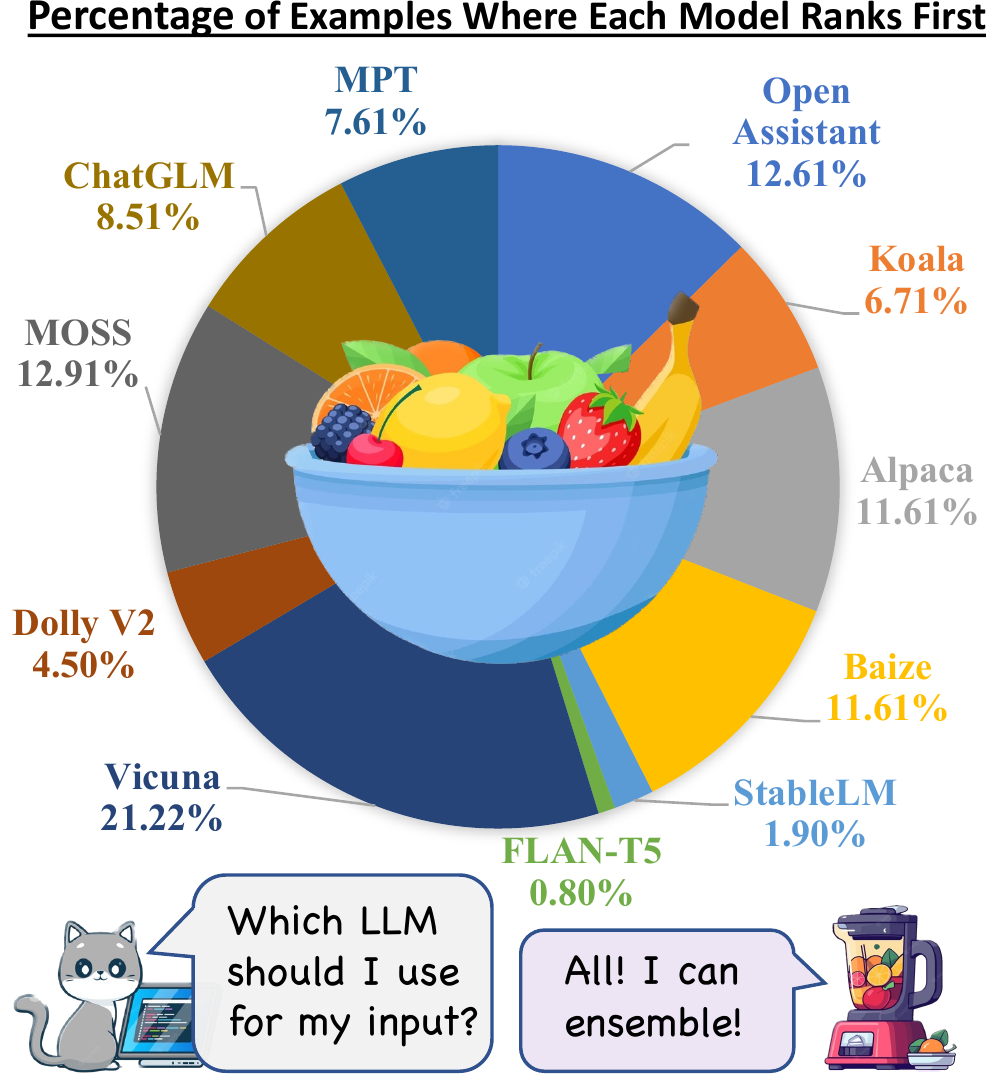}
    \caption{
    \textbf{Motivation of ensembling LLMs.} 
    Based on this pie chart about the percentage of examples where each LLM ranks 1st, we can see that optimal LLMs for different examples can significantly vary. 
    }
    \label{fig:intro}
\end{figure}

The open-source LLMs exhibit diverse strengths and weaknesses due to variations in data, architectures, and hyperparameters, making them complementary to each other. Figure~\ref{fig:intro} illustrates the distribution of best LLMs on 5,000 instructions that we collected. More ranking details can be found in Sec.~\ref{ssec:eval_setup}. Although Vicuna achieves the highest percentage, it ranks first in only 21.22\% of the examples. 
Furthermore, the pie chart suggests that the optimal LLMs for different examples can significantly vary and there is no open-source LLM that dominates the competition. 
Therefore, it is important to dynamically ensemble these LLMs to generate consistently better responses for each input.
Considering the diverse strengths and weaknesses of LLMs, it is crucial to develop an ensembling method that harnesses their complementary potentials, leading to improved robustness, generalization, and accuracy. By combining their unique contributions, we can alleviate biases, errors, and uncertainties in individual LLMs, resulting in outputs better aligned with human preferences.

We introduce \blender \textbf{\methodname{}}, an ensembling framework designed to achieve consistently superior performance by mixing the outputs of multiple LLMs. \methodname{} comprises two modules: \ranker{} and \fuser{}. Initially, \ranker{} compares the outputs from $N$ LLMs, which \fuser{} then fuses to generate the final output from the top $K$ ranked outputs.

Existing approaches~\cite{Ravaut2022SummaRerankerAM, Liu2021SimCLSAS}, including the reward model within InstructGPT~\cite{Ouyang2022TrainingLM}, for ranking outputs $\{y_1, \dots, y_N\}$ from language models (LMs) on a given input $x$ have mostly focused on \textit{individually scoring} each $y_i$ based on $x$, employing encoding modules in the form of $s_i = f_{\phi}(x,y_i)$. Although this list-wise ranking objective can be powerful and efficient when candidate differences are apparent, it may not be as effective when ensembling LLMs. Among the output candidates from LLMs, candidate differences can be quite \textit{subtle}, as they are all produced by very sophisticated models and one may only be marginally better than another. Even for humans, it can be challenging to gauge candidate quality without direct comparison.

As a result, we propose a specialized \textit{pairwise} comparison method, \textbf{\ranker{}} (Sec.~\ref{sec:ranker}), to effectively discern subtle differences between candidate outputs and enhance ranking performance. In particular, we first gather the outputs from $N$ models (e.g., the $N=11$ models in Fig.~\ref{fig:intro}) for each input and subsequently create the $N(N-1)/2$ pairs of their outputs. We jointly encode the input $x$ and the two candidate outputs $y_i$ and $y_j$ as input to a cross-attention encoder (e.g., RoBERTa~\cite{Liu2019RoBERTaAR}), in the form of $f_\phi(x, y_i, y_j)$, to learn and determine which candidate is better.

During the inference stage, we compute a matrix containing logits representing pairwise comparison results. Given this matrix, we can infer a ranking of the $N$ outputs for the given input $x$.
Subsequently, we can employ the top-ranked candidate from \ranker{} for each input as the final result.
Hence, this approach does not rely on a single model for all examples; instead, \ranker{} selects the best model for each example by comprehensively comparing all candidate pairs.

Nonetheless, this approach may constrain the potential to generate even better outputs than the existing candidates.
To investigate this possibility, we introduce the \textbf{\fuser{}} (Sec.~\ref{sec:fuser}) module to fuse the top $K$ of the $N$ ranked candidates and generate an improved output for end-users. Our goal is to capitalize on the strengths of the top $K$ selected candidates while mitigating their weaknesses.

To assess the effectiveness of LLM ensembling methods, we introduce a benchmark dataset called \mixdata \textbf{\dataname{}} (Sec.~\ref{ssec:dataset}). In this dataset, we use $N$=11 popular open-source LLMs to generate $N$ candidates for each input across various existing instruction-following tasks formatted as self-instruct~\cite{Wang2022SelfInstructAL}. The dataset comprises 100k training examples and 5k validation examples for training a candidate ranking module like our \ranker{}, and 5k test examples with oracle comparisons for automatic evaluation.

In Section~\ref{sec:eval}, our empirical results on the \dataname{} benchmark reveal that the \methodname{} framework significantly boosts overall performance by ensembling LLMs. The selections made by \ranker{} outperform any fixed individual LLM models, as indicated by superior performance in both reference-based metrics and GPT-Rank. By leveraging the top selections from \ranker{}, \fuser{} further enhances response quality through effective fusion into the final output.
\methodname{} achieves the highest scores in terms of both conventional metrics (i.e., BERTScore, BARTScore, BLUERT) and ChatGPT-based ranking. The average rank of \methodname{} stands at 3.2 among the 12 methods, which is considerably better than the best LLM's rank of 3.90. Moreover, \methodname{}'s output ranks in the top 3 for 68.59\% of examples, while Viccuna only reaches 52.88\%.
We believe \methodname{} and our findings would benefit both practitioners and researchers for deploying and studying LLMs with ensemble learning.



\begin{figure*}[th!]
    \centering
    \includegraphics[width=1\linewidth]{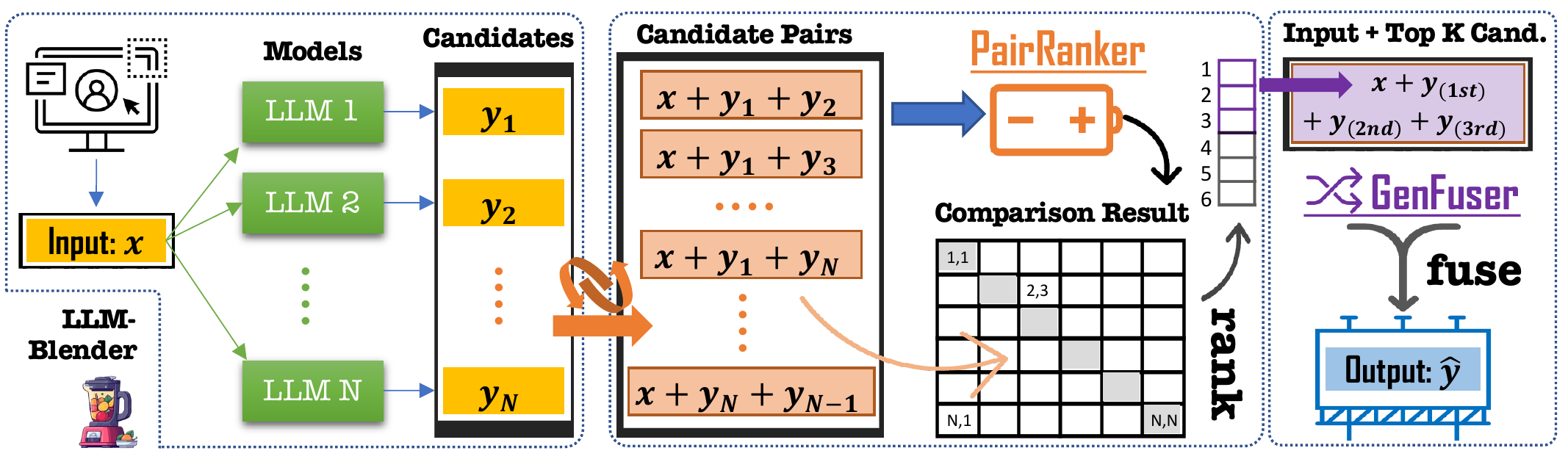}
    \caption{\textbf{The \methodname{} framework.}
    For each input $x$ from users, we employ $N$ different LLMs to get output candidates. Then, we pair all candidates and concatenate them with the input before feeding them to \ranker{}, producing a matrix as comparison results. By aggregating the results in the matrix, we can then rank all candidates and take the top $K$ of them for generative fusion.
    The \fuser{} module concatenates the input $x$ with the $K$ top-ranked candidates as input and generate the final output $\hat{y}$.
    }
    \label{fig:pipeline}
\end{figure*}

\section{Preliminaries}
\label{sec:preliminaries}
We first provide the problem formulation and two common types of ensembling methods. Next, we present the dataset \dataname{} created for training and evaluation purposes. Finally, we give an overview of our framework.

\subsection{Problem Setup}
\label{ssec:problem_setup}
Given an input $x$ and $N$  models, $\{\mathcal{M}_1, \dots, \mathcal{M}_N\}$, we can generate $N$ candidate outputs by processing $x$ with each model. We denote the candidates as $\mathbb{Y}=\{y_1, \dots, y_N\}$. In the training data, we assume there is a ground truth output, $y$, while it remains hidden during evaluation at test time.

In practice, one might choose a fixed model, such as $\mathcal{M}_9$, to infer all unseen examples (i.e., always using $y_9$ as the final output for $x$). This can be reasonable if $\mathcal{M}_9$ demonstrates significantly better overall performance on certain observed examples. However, relying on a pre-selected model may result in sub-optimal performance, as the $N$ models likely possess different strengths and weaknesses in various situations, meaning that the optimal selection for different $x$ values may not always originate from the same model.

Our objective is to develop an ensemble learning method that produces an output $\hat{y}$ for the input $x$, maximizing the similarity $Q(\hat{y}, y; x)$. The $Q$ function can be implemented in various ways, which we will discuss later. We anticipate that this method will yield better overall performance than using a fixed model or randomly selecting a model for $x$. Specifically, given a test set $D_{\text{test}}=\{(x^{(i)},y^{(i)})\}$, we aim to maximize $\sum_i{Q(\hat{y}^{(i)}, y^{(i)}; x^{(i)})}$.


There are two primary approaches for ensembling LLMs: \textit{selection-based} and \textit{generation-based} methods. Selection-based methods compare candidates in the set $\mathbb{Y}$, selecting the top-ranked candidate as the final output $\hat{y}$, which implies that $\hat{y} \in \mathbb{Y}$. Due to the inherent nature of selection and the limited solution space, the performance of selection-based methods is bounded by the $N$ candidates being considered. Conversely, generation-based methods focus on fusing $K$ candidates ($1<K\leq N$) from $\mathbb{Y}$ to produce an unseen response as the final output $\hat{y}$.


\begin{table}[]
    \centering
    \scalebox{0.82}{
    \begin{tabular}{@{} c c c c @{}}
        \toprule
         \makecell[c]{
            \textbf{Sources}
        } & 
        \makecell[c]{
            \textbf{\#Examples}
        } & 
        \makecell[c]{
            \textbf{Source}
        } & 
        \makecell[c]{
            \textbf{I/O Tokens}
        } \\
         \midrule
         Alpaca-{\small{GPT4}} & 22,862 & GPT-4 &  22 / 48 \\
         Dolly-{\small{15K}} & 7,584 & Human & 24 / 53 \\
         GPT4All-{\small{LAION}} & 76,552 & ChatGPT & 18 / 72 \\
         ShareGPT & 3,002 & ChatGPT & 36 / 63 \\
         \midrule
         Total & 110K & Mix & 20 / 66 \\
         \bottomrule
    \end{tabular}
    }
    \caption{\textbf{Statistics of \dataname{}}. It contains 110K examples and we randomly split the dataset into train/dev/test in 100K/5K/5K sizes. }
    \label{tab:llm_dataset_v2}
\end{table}

\subsection{\mixdata \dataname{}: A New Benchmark}
\label{ssec:dataset}
We introduce a new dataset, \dataname{}, to benchmark ensemble models for LLMs in instruction-following tasks. We collect a large-scale set of instruction examples primarily from four sources, as shown in Table~\ref{tab:llm_dataset_v2}. After curating and processing this open-source data, we sample 100k examples for training, 5k for validation, and 5k for testing. We then run $N=11$ popular open-source LLMs, including Vicuna, OpenAssistant, Alpaca, MPT, and others (see Table~\ref{tab:llm_results} and Figure~\ref{fig:intro}), on these 110k examples.

To obtain the oracle ranking of candidates, we design comparative prompts for ChatGPT to evaluate all candidate pairs. 
Specifically, for each example, we prepare 55 pairs of candidates ($11\times10/2$). For each pair, we ask ChatGPT to judge the better candidate (or declare a tie).
The prompt template can be found in the appendix. For the training and validation sets, we provide the results based on conventional  metrics like BERTScore, BLEURT, and BARTScore. In that case, we use function $Q(y_i, y)$ to estimate a candidate $y_i$'s quality according to its similarity to the ground truth $y$.

\subsection{\blender \methodname{}: A Novel Framework}
\label{ssec:overview}
We propose a rank-and-fuse pipeline framework, \methodname{}, for ensembling LLMs, as illustrated in Figure~\ref{fig:pipeline}. This framework consists of two main components: a pairwise ranking module, \ranker{} (Section~\ref{sec:ranker}), and a fusion module, \fuser{} (Section~\ref{sec:fuser}). The \ranker{} module learns to compare all pairs of candidates for each input and subsequently rank the list of candidates. We then select the top $K=3$ ranked candidates, concatenate them with the input $x$, and construct the input sequence for the \fuser{} module. The \fuser{} module, a seq2seq LM, ultimately generates the final output to serve users. 

\section{\textsc{PairRanker}: Pairwise Ranking}
\label{sec:ranker}
\begin{figure*}[t]
    \centering
    \includegraphics[scale=0.8]{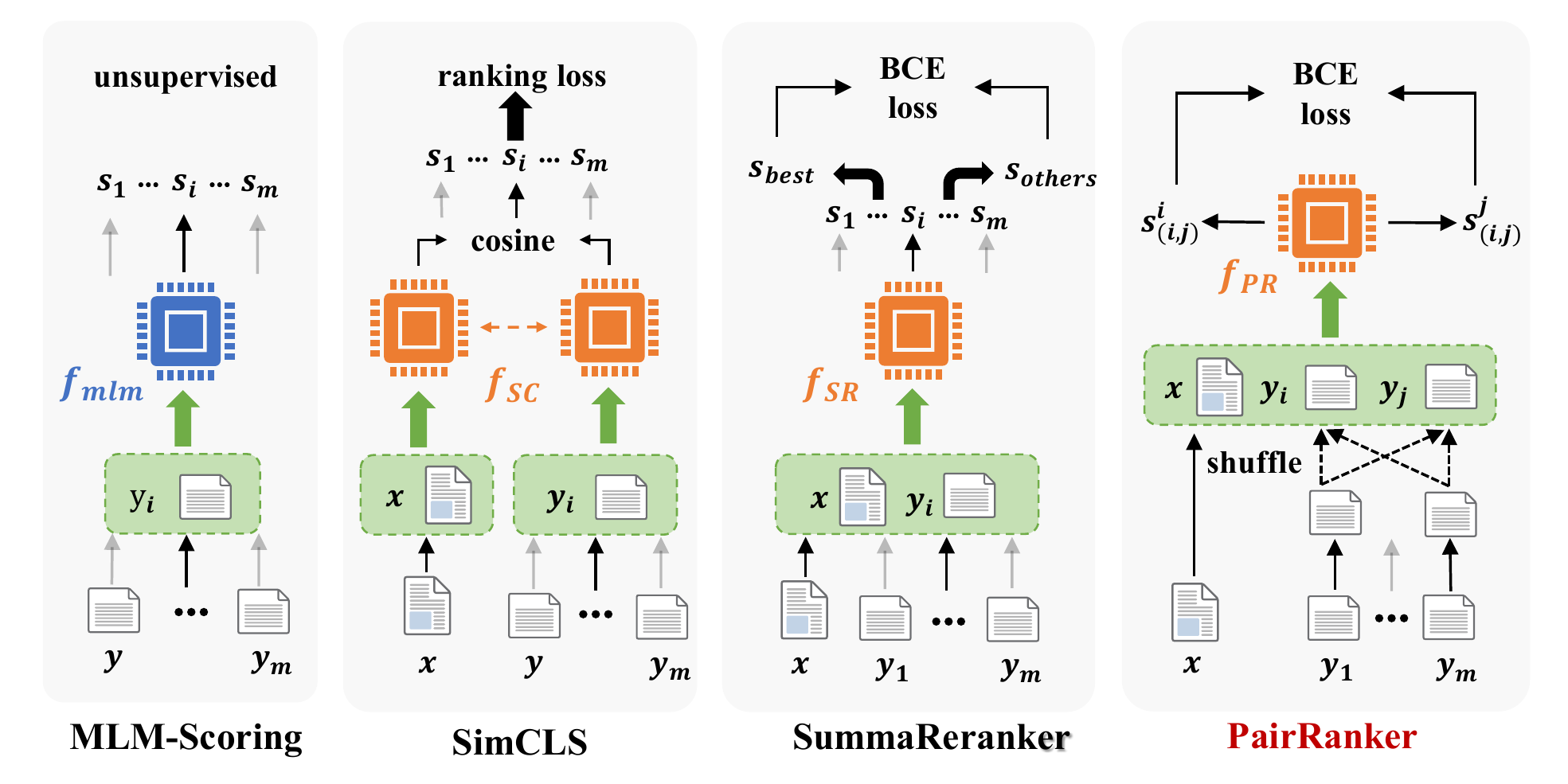}
    \caption{\textbf{The architectures of typical reranking methods.} $x$ is an input and $y_i$ is a certain candidate, and its score is $s_i$. MLM-Scoring is an unsupervised method that uses an external masked LM to score a candidate; SimCLS uses the same encoder to encode $x$ and each candidate $y_i$; SummaReranker instead employs a cross-encoder to encode both $x$ and $y_i$ at the same time; \ranker{} encodes a pair of candidates ($y_i$, $y_j$) at the same time for pairwisely scoring them, and the final score of each candidate is produced as shown in Fig.~\ref{fig:matrix}.
    }
    \label{fig:model_archs}
\end{figure*}

In this section, we introduce three baseline methods for ranking the candidates in $\mathbb{Y}$ in Sec.~\ref{ssec:individual_scoring} and present the proposed \ranker{} method.

\subsection{Baseline Methods}
\label{ssec:individual_scoring}
Previous reranking methods primarily focus on computing the score $s_i=f_\phi(x, y_i)$ for each candidate $y_i \in \mathbb{Y}$ independently, where $s_i$ is solely determined by $y_i$. Notably, the reward model in instruction tuning for GPT-3.5~\cite{Ouyang2022TrainingLM} also belongs to this category. Figure~\ref{fig:model_archs} illustrates these baseline methods, which are further detailed in the following paragraphs.

\textbf{MLM-Scoring}~\citep{mlmscoring} assesses the quality of a candidate by calculating its pseudo-log-likelihood, which is obtained by masking tokens one by one and computing the log-likelihood for the masked token using masked LMs (e.g., BERT). Given a candidate $y_i$ as a sequence of words $\textbf{W}=\{w_1, ..., w_{|\textbf{W}|}\}$, the pseudo-log-likelihood is:
$s_i = \sum_{t=1}^{|\textbf{W}|}\log{P(w_t|\textbf{W}_{\backslash t})}$. This unsupervised method is effective for reranking outputs in NLG tasks such as machine translation and speech recognition.

\textbf{SimCLS}~\citep{Liu2021SimCLSAS} encodes the input $x$ and each generated candidate $y_i\in \mathbb{Y}$ using the same encoder $H$, resulting in $H(x)$ and $H(y_i)$. The cosine similarity between them, $s_i=\cos{(H(x), H(y_i))}$, serves as the predicted score, as $H(x)$ and $H(y_i)$ share the same embedding space induced by the language encoder. In training, marginal ranking loss is used to optimize $H$.

\textbf{SummaReranker}~\citep{Ravaut2022SummaRerankerAM} concatenates the input $x$ and each candidate $y_i$, using a cross-attention encoder to learn ranking. Specifically, they employ $H([x;y_i])$ to predict the score $s_i$, where $H$ is a Transformer model. In the training stage, binary cross-entropy (BCE) loss is employed to differentiate the best candidate from the others.

\paragraph{Limitations.}
Despite using contrastive loss in training, these methods rely on individual scoring for inference. The encoders have not been exposed to pairs of candidates for direct comparison learning. We argue that such pointwise ranking methods may be insufficient for selecting the best candidates in the context of LLMs and instruction-following tasks. One reason is that the quality of LLM outputs is generally high when the chosen LLMs are popular and competitive. Moreover, the responses for instruction tasks can be quite open-ended, unlike summarization tasks. Therefore, merely examining individual candidates may not yield a reliable score. This issue becomes more prominent for shorter responses, where sequences may differ by only a few words but vary significantly in helpfulness, harmfulness, and fairness. Given these limitations, we contend that individual scoring approaches may fail to capture crucial nuances.

\subsection{Pairwise Comparisons}
\label{ssec:pairwise_comparison}

In order to address the limitations of pointwise ranking, we aim to train a ranker $f$ with parameter $\phi$ that can compare a pair of output candidates by encoding them together with the input text. Our ranker module should focus on learning to capture the differences between the two candidates and prefer the ones of higher quality.

Given a pair of candidates $y_i, y_j$, we obtain their pair-specific scores: $s_{(i,j)}^i$ and $s_{(i,j)}^j$. We denote the model's confidence in thinking $y_i$ is better than $y_j$ as $s_{ij} = s_{(i,j)}^i - s_{(i,j)}^j$. We can use these scores for all pairs induced from $\mathbb{Y}$ to infer the final ranking. To learn this ability, we concatenate the input $x$ and the two candidates to form a sequence $[x;y_i;y_j]$ and feed it into a cross-attention Transformer to get the features: $f_{\phi}([x;y_i;y_j])$ for modeling $s_{ij}$.

We assume multiple $Q$ functions to optimize for, such as BERTScore, BARTScore, etc., and consider the learning problem as a multi-task classification problem:
\begin{align*}
    \mathcal{L}_Q = -z_i\log{\sigma(s_{(i,j)}^i)} - z_j\log{\sigma(s_{(i,j)}^j)},
\end{align*}
where $\sigma$ denotes the sigmoid function and 
\begin{align*}
    (z_i, z_j) = 
    \left\{
        \begin{aligned}
            (1, 0) &, &Q(y_i, y) \ge Q(y_j,y) \\
            (0, 1) &, &Q(y_i, y) < Q(y_j,y)
        \end{aligned}
    \right. .
\end{align*}
For optimizing towards multiple $Q$, we take the average as the final multi-objective loss: $\mathcal{L}= \sum \mathcal{L}_Q $.

\subsection{\textsc{PairRanker} Architecture}
\label{sec:architecture}

We discuss the concrete designs for the \ranker{} module in this subsection.

\paragraph{Encoding.}
We employ Transformer layers to encode an input and a pair of candidates, enabling the attentions to capture the difference between candidates in the context of the input. We concatenate the three segments sequentially and form a single input sequence with special tokens as separators: \texttt{<source>}, \texttt{<candidate1>}, and \texttt{<candidate2>}. The resulting input sequences to Transformers are in the form of ``\texttt{<s><source> $x$} \texttt{</s>} \texttt{<candidate1>} $y_i$ \texttt{</s>} \texttt{<candidate2> $y_j$ </s>}'', where $x$ is the text of a source input and $y_i$ and $y_j$ are the text of two output candidates. The embeddings of special tokens $\texttt{<source>}$, $\texttt{<candidate1>}$, and $\texttt{<candidate2>}$ are used as the representations of $x$, $y_i$, and $y_j$ respectively, denoted as $\mathbf{x}$, $\mathbf{y_i}$, $\mathbf{y_j}$.

\paragraph{Training.}
To determine the scores for the two candidates, we concatenate the embeddings of $\mathbf{x}$ with $\mathbf{y_i}$ and $\mathbf{y_j}$ respectively, and pass them through a single-head layer, which is a multi-layer perceptron with the final layer's dimension equal to the number of $Q$ functions to be optimized. Each value within this dimension represents a computed $Q$ score for a specific $Q$ function. We derive the final score $s_{(i,j)}^i$ or $s_{(i,j)}^j$ for the candidate by averaging these $Q$ scores. Since there are $O(N^2)$ unique pair combinations, we apply an effective sub-sampling strategy during the training stage to ensure learning efficiency.

During training, we randomly select some combinations from the candidate pool $\mathbb{Y}^2$, instead of all the $N(N-1)/2$ pairs. We also compare the target text with other candidates by extending the candidate pool by mixing the ground truth $y$ into $\mathbb{Y}$. In practice, we found that using 5 pairs per input is sufficient for obtaining decent results.

Due to the position embeddings of the language model, the order of the candidates in a pair $(x,y_i,y_j)$ matters, as the comparison result of  $(x, y_i,y_j)$ and $(x,y_j,y_i)$ might not be consistent. Thus, we shuffle the order of candidates within each training pair so that the model learns to be consistent with itself.

 \paragraph{Inference.}
During the inference stage, we obtain scores $s_{ij}$ for each pair of candidates $(y_i, y_j)\in \mathbb{Y}^2$. After $N(N-1)$ iterations, we obtain a matrix $\mathbf{M}$, where $\mathbf{M}_i^j=s_{ij}$ represents the \textit{confidence} that $y_i$ is better than $y_j$. To identify the best candidate based on $\mathbf{M}$, we introduce three aggregation functions for determining the final ranking of $\mathbb{Y}$.

We propose two scoring methods, \texttt{MaxLogits} and \texttt{MaxWins}, which utilize all elements in the matrix. Let $\mathbf{M}_i^*$ and $\mathbf{M}_*^j$ denote the $i$-th row and $j$-th column of the matrix as vectors. For each candidate $y_i$, its \texttt{MaxLogits} score is defined as $s_i=\sum{(\mathbf{M}_i^*-\mathbf{M}_*^i)}$, while its \texttt{MaxWins} score is defined as $s_i=|\{s_{ij}\in \mathbf{M}_i^*|s_{ij}>0\}| + |\{s_{ji}\in \mathbf{M}_*^i|s_{ji}<0\}|$, where $||$ denotes the set size.

In essence, \texttt{MaxLogits} computes the confidence that $y_i$ is superior to all other candidates, whereas \texttt{MaxWins} calculates the number of victories in comparisons with other candidates.

However, these two methods necessitate $O(N^2)$ iterations for $N$ candidates, which can be computationally burdensome. Thus, we propose a more efficient aggregation method, performing \textit{a single bubble sort run} with pairwise comparisons to select the best candidate. We first shuffle the order of candidates in $\mathbb{Y}$ to obtain a default order, and initialize the best candidate index $k$ to 1. We iteratively update the best candidate index as follows:
\begin{align*}
    \small 
    k = \left\{
        \begin{aligned}
            k&, & \mathbf{M}_{k}^{i} - \mathbf{M}_{i}^{k} > 0\\ 
            i&, & \mathbf{M}_{i}^{k} - \mathbf{M}_{k}^{i} > 0
        \end{aligned}
    \right. .
\end{align*}
After $N-1$ comparisons, we select $y_k$ as the best candidate. This method reduces the inference time complexity from $O(N^2)$ to $O(N)$, aligning with previous pointwise methods.

Regardless of the aggregation method, we can rank all candidates in $\mathbb{Y}$. Our experiments (shown in the appendix) reveal that \texttt{MaxLogits} yields the best performance, so we use \texttt{MaxLogits} as the default aggregator for \ranker{}.
\begin{figure}
    \centering
    \includegraphics[width=1\linewidth]{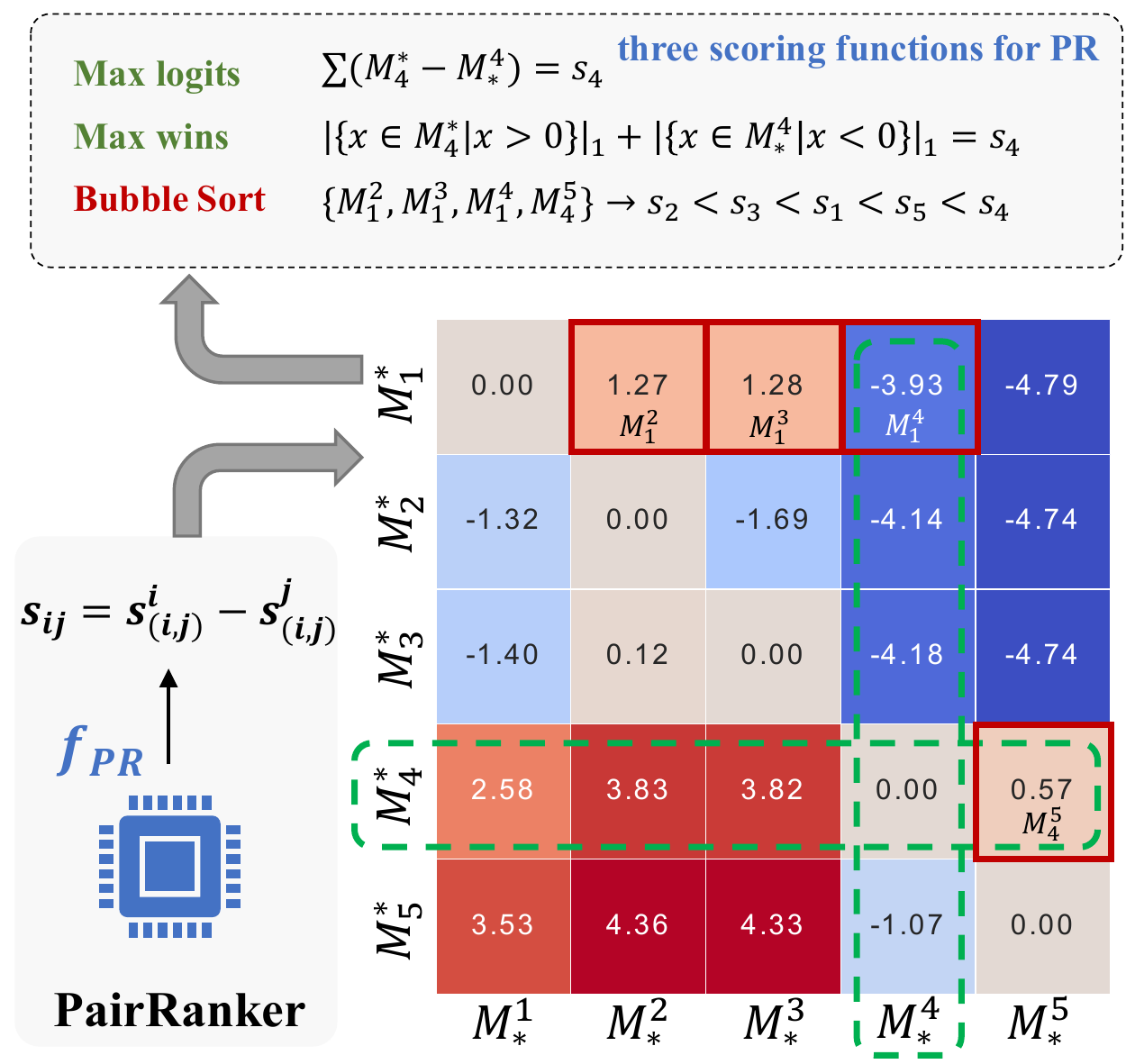}
    \caption{\textbf{Aggregation methods for \ranker{}.}  \vspace{-1em}}
    \label{fig:matrix}
\end{figure}




\section{\textsc{GenFuser}: Generative Fusion}
\label{sec:fuser}

The effectiveness of \ranker{} is constrained by the quality of selections from the candidate pool $\mathbb{Y}$. We hypothesize that by merging multiple top-ranked candidates, we can overcome this constraint. As these top candidates often showcase complementary strengths and weaknesses, it is plausible to generate a superior response by combining their advantages while mitigating their shortcomings. Our objective is to devise a generative model that takes input $x$ and $K$ top-ranked candidates $\{y_1, ..., y_K\} \subset \mathbb{Y}$ (e.g., $K=3$) and produces an improved output $\hat{y}$ as the final response.

To accomplish this, we present \fuser{}, a seq2seq approach for fusing a set of candidates conditioned on the input instruction to generate an enhanced output. Specifically, we concatenate the input and $K$ candidates sequentially using separator tokens, such as \texttt{<extra\_id\_$i$>}, and fine-tune a T5-like model to learn to generate $y$. In practice, we employ Flan-T5-XL~\cite{Chung2022ScalingIL}, which has 3b parameters, due to its superior performance and relatively smaller size.

\begin{table*}[htb]
    \centering
    \scalebox{0.64}{
    \begin{tabular}{@{}c c | c c  c ||c  c c c @{}}
        \toprule
        \textbf{Category} & \textbf{Methods}
         & \textbf{BERTScore}$\uparrow$ & \textbf{BARTScore}$\uparrow$ & \textbf{BLEURT}$\uparrow$ & \textbf{GPT-Rank}$\downarrow$ & 
         $\ge$ \textbf{Vic}(\%)$\uparrow$ & $\ge$ \textbf{OA}(\%)$\uparrow$ & Top-3(\%)$\uparrow$\\
                          \midrule
\multirow{11}{*}{\textbf{LLMs}}    & \textbf{O}pen \textbf{A}ssistant~\cite{oasst} & \textbf{74.68} & -3.45          & \textbf{-0.39} & \textbf{3.90} & \textbf{62.78}  & N/A             & 51.98           \\
                          & \textbf{Vic}una \cite{vicuna2023}    & 69.60          & \textbf{-3.44} & -0.61          & 4.13          & N/A             & \textbf{64.77}  & \textbf{52.88}  \\
                          & Alpaca~\cite{alpaca}                 & 71.46          & -3.57          & -0.53          & 4.62          & 56.70           & 61.35           & 44.46           \\
                          & Baize~\cite{xu2023baize}             & 65.57          & -3.53          & -0.66          & 4.86          & 52.76           & 56.40           & 38.80           \\
                          & MOSS~\cite{moss}                     & 64.85          & -3.65          & -0.73          & 5.09          & 51.62           & 51.79           & 38.27           \\
                          & ChatGLM~\cite{du2022glm}             & {70.38} & {-3.52} & {-0.62} & 5.63          & 44.04           & 45.67           & 28.78           \\
                          & Koala~\cite{koala_blogpost_2023}     & 63.96          & -3.85          & -0.84          & 6.76          & 39.93           & 39.01           & 22.55           \\
                          & Dolly V2~\cite{dollyV2}              & 62.26          & -3.83          & -0.87          & 6.90          & 33.33           & 31.44           & 16.45           \\
                          & Mosaic MPT~\cite{MosaicML2023Introducing}     & 63.21          & -3.72          & -0.82          & 7.19          & 30.87           & 30.16           & 16.24           \\
                          & StableLM~\cite{stablelm}             & {62.47} & {-4.12} & {-0.98} & 8.71          & 21.55           & 19.87           & 7.96            \\
                          & Flan-T5~\cite{Chung2022ScalingIL}    & 64.92          & -4.57          & -1.23          & 8.81          & 23.89           & 19.93           & 5.32            \\
                          \midrule
\multirow{4}{*}{\textbf{Analysis}} & Oracle (BERTScore)                   & \textbf{77.67} & -3.17          & -0.27          & 3.88          & 54.41           & 38.84           & 53.49           \\
                          & Oracle (BLEURT)                      & 75.02          & -3.15          & \textbf{-0.15} & 3.77          & 55.61           & 45.80           & 55.36           \\
                          & Oracle (BARTScore)                   & 73.23          & \textbf{-2.87} & -0.38          & 3.69          & 50.32           & 57.01           & 57.33           \\
                          & Oracle (\texttt{GPT-Rank})           & 70.32          & -3.33          & -0.51          & \textbf{1.00} & \textbf{100.00} & \textbf{100.00} & \textbf{100.00} \\
                          \midrule
\multirow{5}{*}{\textbf{Rankers}}  & Random                               & 66.36          & -3.76          & -0.77          & 6.14          & 37.75           & 36.91           & 29.05           \\
                          & MLM-Scoring                          & 64.77          & -4.03          & -0.88          & 7.00          & 33.87           & 30.39           & 21.46           \\
                          & SimCLS                               & \textbf{73.14} & -3.22          & -0.38          & 3.50          & 52.11           & 49.93           & 60.72           \\
                          & SummaReranker                        & 71.60          & -3.25          & -0.41          & 3.66          & \textbf{55.63}  & 48.46           & 57.54           \\
                          & \textbf{PairRanker}                  & 72.97          & \textbf{-3.14} & \textbf{-0.37} & \textbf{3.20} & 54.76           & \textbf{57.79}  & \textbf{65.12}  \\
                          \midrule
\textbf{\methodname{}}    & \textbf{PR ($K=3$) + GF}             & \textbf{79.09} & \textbf{-3.02} & \textbf{-0.17} & \textbf{3.01} & \textbf{70.73}  & \textbf{77.72}  & \textbf{68.59}  \\
                          \bottomrule 
                          \bottomrule
    \end{tabular}
    }
    \caption{
       Empirical results on \dataname{}. GPT-Rank are the most important metric.
    }
    \label{tab:llm_results}
\end{table*}

\section{Evaluation}
\label{sec:eval}

\subsection{Setup}
\label{ssec:eval_setup}
We use \dataname{} (Sec.~\ref{ssec:dataset}) to conduct evaluation, and more results are in the appendix.

\paragraph{NLG metrics.}
We employ two types of evaluation metrics (i.e., $Q$ ).
The first group is conventional automatic metrics for NLG tasks: BERTScore~\cite{Zhang2020BERTScoreET}, BLEURT~\cite{Sellam2020BLEURTLR}, and BARTScore~\cite{Yuan2021BARTScoreEG}. 

\paragraph{GPT-Rank.}
The second is based on prompting ChatGPT for pairwise comparisions on all candidates and decide their rank by the number of wins (i.e., \texttt{MaxWins} aggregation). 
We name this GPT-based ranking metric with \texttt{GPT-Rank}.

\paragraph{Model training.}
We use the DeBERTa~\cite{He2021DeBERTaV3ID} (400m) as the backbone for \ranker{}, and \fuser{} is based on Flan-T5-XL (3b). 
According to our ablation studies, we choose to use BARTScore for its superior correlation with \texttt{GPT-Rank} as shown in ~\ref{pacorrelation}.

\subsection{Main results}
\label{ssec:main_results}
In Table~\ref{tab:llm_results}, we present the overall performance of $N$=11 LLMs as well as other methods on \dataname{}. 
In addition to the three auto metrics and GPT-Rank, we also show the percentage of examples where each method can produce outputs that are \textit{better than or same good} as the two top LLMs, namely OpenAssistant ($\ge$OA) and Vicuna ($\ge$Vic), in terms of \texttt{GPT-Rank}.

\paragraph{LLMs have diverse strengths and weakness.}
The table presents the LLMs in a sorted order based on their average rank as determined by ChatGPT (GPT-Rank). Among these models, Open Assistant, Vicuna, and Alpaca are the top-3 performers. Following them, three renowned LLMs, namely Baize, Moss, and ChatGLM, which have been fine-tuned using both Chinese and English instruction data, also exhibit impressive performance on \dataname{}. 
Conversely, Mosaic MPT, StableLM, and Flan-T5 rank at the bottom-3 in the evaluation. Nevertheless, the average \texttt{GPT-Rank} of top/bottom models maintain a noticeable distance from the first/last position (1 or 11), highlighting the importance of ensembling LLMs. 


\paragraph{Top LLMs are not always good.}
It is evident that although OA and Vic perform remarkably well, there is still a substantial percentage of examples where other LLMs are considered to outperform them. For instance, despite Koala having an average \texttt{GPT-Rank} of $6.76$, approximately $40\%$ of the examples demonstrate that Koala produces responses that are better or equally as good as both OA and Vic. This further emphasizes the significance of employing our \methodname{} framework for ranking and fusion purposes.

\paragraph{NLG Metrics.}
Moreover, we conduct a comprehensive analysis of the performance of oracle (top-1) selections based on each of the metrics themselves. The findings demonstrate that these selections also exhibit favorable performance across other metrics as well. For example, the oracle selections derived from \textbf{GPT-Rank} achieve a BARTScore of $-3.33$, surpassing that of OA ($-3.45$). Conversely, the oracle selections of BARTScore yield $3.69$ in \texttt{GPT-Rank}, also significantly outperforming OA ($3.90$). This observation substantiates the rationality of using BARTScore to provide supervision for \ranker{}, which is also suggested by Table~\ref{tab:rank_correlation}.


\paragraph{\ranker{} outperforms other rankers.}
MLM-Scoring fails to outperform even random selection, highlighting the limitations of its unsupervised paradigm. On the contrary, SimCLS, SummaReranker, and \ranker{} exhibit superior performance compared to the best model (OA) across BARTScore  and \texttt{GPT-Rank}.  
Notably, the average \texttt{GPT-rank} of the responses selected by \ranker{} ($3.20$) significantly outperforms the best model by $0.70$ (a 18\% relative performance gain) and also all other rankers. Moreover, it achieves impressive results in metrics such as BARTScore ($-3.14$) with a substantial advantage.
\ranker{}'s selections are better than or equal to Vic/OA on $54.76\%$/$57.79\%$ examples respectively, and ranks in top 3 for 65.12\% examples.  

\paragraph{\methodname{} is the best.}
We use top-3 selections from the \ranker{} and feed them as candidates for \fuser{}. 
Based on this integration, \methodname{} demonstrates remarkable capabilities as expected. 
In terms of \texttt{GPT-Rank}, it achieves $3.01$, surpassing both the best model OA ($3.90$) by a significant margin. The scores for BERTScore ($79.09$), BARTScore ($-3.02$), and BELURT ($-0.17$) all exceed the best model by $4.41$, $0.43$, and $0.22$ respectively, showcasing substantial advantages.
Moreover, \methodname{} also performs well in surpassing the top two models, Vic ($70.73$) and OA ($77.72$), thereby complementing the weaknesses of \ranker{}.

\paragraph{Ranking correlation.}
\label{pacorrelation}
\begin{table}[]
    \centering
    \scalebox{0.7}{
    \begin{tabular}{@{} c c c c @{}}
        \toprule
        \textbf{Ranking Methods} & \makecell[c]{Pearson\\Correlation $\uparrow$}  & \makecell[c]{Spearman's\\Correlation $\uparrow$ } & \makecell[c]{Spearman's\\Footrule $\downarrow$} \\
        \midrule
        Random              & 0.00           & 0.00           & 48.27          \\
        BLEU                & 28.70          & 26.92          & 33.57          \\
        Rouge2              & 29.17          & 27.77          & 32.96          \\
        BERTScore           & 32.25          & 30.33          & 33.34          \\
        BLEURT              & 34.14          & 32.31          & 32.17          \\
        BARTScore           & \textbf{38.49} & \textbf{36.76} & \textbf{30.93} \\
        \midrule
        MLM-Scoring         & -0.02          & -0.01          & 47.16          \\
        SimCLS              & 39.89          & 38.13          & 29.32          \\
        SummaReranker       & 41.13          & 39.10          & 29.69          \\
        \textbf{PairRanker} & \textbf{46.98} & \textbf{44.98} & \textbf{27.52} \\
        \bottomrule
    \end{tabular}
    }
    \caption{The correlation between each ranking method and oracle ranking (GPT-Rank).}
    \label{tab:rank_correlation}
\end{table}

In addition to focusing solely on the top-1 selection of each ranker, we present a comprehensive analysis of the overall rank correlation among all the candidates with \texttt{GPT-Rank} (see Table \ref{tab:rank_correlation}). The correlation metrics used here include the Pearson Correlation Coefficient, Spearman's Correlation, and Spearman's Footrule distance\cite{Diaconis1977SpearmansFA}.

It turns out that BARTScore gets the highest correlation with \texttt{GPT-Rank} against other metrics, which suggests we use BARTScore to provide supervision for training.
For rankers, MLM-Scoring still falls short of outperforming random permutations. On the other side, SummaReranker demonstrates better correlation in terms of the Pearson Correlation ($41.13$) and Spearman's Correlation ($39.10$), while SimCLS gets a better Spearman's Footrule distance ($29.32$)
Notably, \ranker{} achieves the highest correlation with \texttt{GPT-Rank} across all correlation types, which is even way better than the BARTScore.

\paragraph{More analysis.}
We leave many other ablation studies and analyses in Appendix, where we apply \ranker{} to the three typical natural language generation (NLG) tasks: summarization (CNN/DM), machine translation (WMT18-zh-en), and constrained text generation (CommonGen).
We find that \ranker{} still outperforms other methods by a large margin in the context of using a single same base model to decode $N$ candidates (with different algorithms).
We also show that \texttt{MaxLogits} is much better than \texttt{MaxWins} and the bubble sort method is very cost-effective if the inference efficiency is a big concern. 

\section{Related Work}
\label{sec:related}

\paragraph{LLM evaluation}
As open-source large language models (LLMs) continue to flourish and demonstrate remarkable competitiveness across various natural language generation (NLG) tasks, assessing the capabilities of LLMs has become an exceedingly challenging endeavor.
To address this issue, \citet{zheng2023judging} pioneered the creation of a chatbot arena, enabling users to provide pairwise evaluations of responses generated by two randomly selected LLMs. Based on these evaluations, they established an LLM Elo rating leaderboard.
In a similar vein, \citet{zeno_chatbot} conducted an evaluation study on a customer service dataset, leveraging automated metrics such as BERTScore and ChrF~\cite{chrf}. This approach yielded similar LLM ranking results.
Instead of relying solely on human evaluation, \citet{PandaLM} developed a fine-tuned model called PandaLM to compare responses generated by different LLMs. 
AlpacaFarm~\cite{Dubois2023AlpacaFarmAS} also aims to evaluate LLMs with pairwise feedback.

\paragraph{Pairwise ranking}
Pairwise ranking, known for its long-standing effectiveness, has demonstrated exceptional performance across a wide array of NLP tasks \cite{Jamieson2011ActiveRU}. Notably, Ranknet~\cite{Burges2005LearningTR} and LambdaRank~\cite{Burges2010FromRT} have emerged as powerful techniques for various ranking problems. Furthermore, within the renowned RLHF procedure\cite{Ouyang2022TrainingLM}, these methods incorporate pairwise training of their reward model based on OPT. However, these approaches still compute scores individually and solely undergo pairwise training at the loss level. In contrast, our proposed \textsc{PairReranker} not only employs pairwise training but also utilizes the attention mechanism for pairwise inference during the inference stage. We posit that this approach better captures the subtleties between candidates and yields superior results, as demonstrated in Section~\ref{ssec:main_results}.

\paragraph{Ensemble learning}
Ensemble learning is a widely employed technique to enhance a model's capabilities by leveraging multiple weaker models~\cite{Sagi2018EnsembleLA,Anio2019EnsembleAF}. Typically, ensemble learning is performed either by considering model weights or by combining diverse outputs. 
Mix-of-Experts (MoE) is a type of ensemble approach that combines the predictions of multiple specialized sub-models to improve overall performance. It has been successfully applied in various domains, such as natural language processing and computer vision~\cite{Jacobs1991AdaptiveMO,Shazeer2017OutrageouslyLN}.
As for fusing multiple candidates, ~\citet{Izacard2020LeveragingPR} introduced a  framework named Fusion-in-Decoder (FiD) to improve the quality of question answering by fusing retrieved text. Building upon FiD, \citet{Ravaut2022TowardsSC} further investigated the effectiveness of fusion in the context of text summarization. However, they neglected to incorporate a selection process prior to feeding the candidates into the fusion module, resulting in only moderate improvements. In contrast, our proposed approach, referred to as \methodname{}, initially utilizes the \ranker{} algorithm to filter out candidates of poor quality. Subsequently, fusion is performed exclusively on the top-ranked candidates, leading to superior performance.

\section{Conclusion \& Future Directions}
\label{sec:conclusion}

In this paper, we formulated the motivation to exploit the diverse strengths and weaknesses of open-source large language models (LLMs), aiming to create an ensembling framework that leverages their complementary capabilities to generate consistently superior results on various instruction-following tasks. By dynamically ensembling LLMs, we aimed to reduce biases, errors, and uncertainties in individual models, yielding outputs better aligned with human feedback.

Our major contributions are as follows:
\begin{itemize}
    \item \textbf{A new framework:}\blender \textbf{\methodname{}} is a post-hoc ensemble learning method for ranking and fusing the outputs from multiple LLMs. 
    It is composed of two modules: \ranker{} and \fuser{}, and both are straightforward yet effective.
    \item \textbf{A new dataset:}\mixdata \textbf{\dataname{}} is a benchmark dataset, created for training and evaluating LLM ensembling methods on instruction-following tasks.
    \item \textbf{Promising results:} We show that our method can significantly improve the overall results on various metrics, and our findings indicates that this direction is promising for both research community and practitioners.
    \item \textbf{Toolkit:} By open-sourcing our framework, we aim to make it easier for others to leverage our approach, enabling the development of more advanced AI systems that achieve robustness, generalization, and enhanced accuracy in a wide variety of tasks. 
\end{itemize}

\paragraph{Future directions.}
Potential future directions include extending the \methodname{} framework to more types of models or even non-text modalities, developing more sophisticated ranking and fusion techniques, and investigating the transferability of our ensembling approach to other domains and tasks. Additionally, exploring ways to minimize computational overhead and incorporating active learning strategies for rapid adaptation to new specialized domains and data sources represent fruitful areas for further research. Overall, our work underscores the value of combining the unique contributions of multiple models.

\section*{*Limitations}
\paragraph{Efficiency.}
To get the optimal performance from \ranker{}, one may need to call the model $O(n^2)$ times for getting the full matrix, thus resulting in a much less efficient solution. 
We attempted to resolve this limitation by proposing to use multiple rounds of bubble sort methods to reduce the number of inferences needed, and we find it works pretty well.
We also want to argue that although the number of inferences can be large for obtaining the best performance with \ranker{}, those inferences can be executed in parallel because they are totally independent.  

\paragraph{Human evaluation.}
We agree that automatic metrics have limitations. Human evaluation could provide us with more reliable and comprehensive evaluation results. However, due to the number of models as well as the amounts of generation candidates, we cannot afford large-scale human evaluation. We argue that our use of ChatGPT for evaluation is a good alternative, according to recent studies. 
Also, we would like to highlight that we show the ground truths when using ChatGPT to do pairwise comparisions, which is quite informative than the common practice.


 
\section*{*Ethical Statement}
This work fully complies with the ACL Ethics Policy.
We declare that there are no ethical issues in this paper, to the best of our knowledge.

\section*{Acknowledgements}
We thank members of the INK lab at USC and the Mosaic team at AI2 for valuable feedback on this project.
Xiang is supported in part by the Office of the Director of National Intelligence (ODNI), Intelligence Advanced Research Projects Activity (IARPA), via the HIATUS Program contract \#2022-22072200006, the DARPA MCS program under Contract No. N660011924033, the Defense Advanced Research Projects Agency with award W911NF-19-20271, NSF IIS 2048211, and gift awards from Google and Amazon. 
Yuchen's research was also supported by the Allen Institute for AI (AI2).
The views and conclusions contained herein are those of the authors and should not be interpreted as necessarily representing the official policies, either expressed or implied, of ODNI, IARPA, or the U.S. Government.


\bibliography{custom}
\bibliographystyle{acl_natbib}

\clearpage
\appendix
\appendix

\begin{tcolorbox}[colframe=black!80, colback=white, sharp corners, boxrule=1pt, arc=5pt, rounded corners, left=1pt, right=1pt, top=2pt, bottom=2pt]
\centering
\large \textbf{Appendix}
\end{tcolorbox}

\section{Implementation Details}
\label{sec:training_details}

\paragraph{\ranker{}}
We train our ranker for 5 epochs. We use the Adafactor optimizer~\cite{Shazeer2018AdafactorAL}, with the maximum learning rate being 1e-5. The warm-up ratio is 5\% with a linear learning rate scheduler. Our training batch size is 64. The training finishes on a single RTX 8000 GPU in two days. The backbone of \ranker{} is Deberta-v3-large~\cite{He2021DeBERTaV3ID}. 
Unlike the mixture-of-experts layer used in the work of \citet{Ravaut2022SummaRerankerAM}, we employ a five-layer multi-layer perceptron (MLP) with the hyperbolic tangent activation function. The output dimension of the final layer is equal to the number of different metrics.
In practice, we have tried different special embedding combinations, such as only feeding \texttt{<candidate1>} and \texttt{<candidate2>}, mean pooling representation, and so on. And Finally, we found that concatenating \texttt{<source>} with \texttt{<candidate1>}, and \texttt{<source>} with \texttt{<candidate2>} respectively achieves the best performance.
We also tried different loss types, like MSE losses and ranking losses. And we find BCE is simply good enough.

\paragraph{\fuser{}} 
We use Flan-T5-large and Flan-T5-xl (3b) to train the top-3 bart-score selections as input and then apply it with \ranker{}'s top 3 selections for inference. 
We find Flan-t5-3b performs much better than the large version, while flan-t5-xxl has marginal improvements yet being much larger and longer to train.

\section{Conventional Tasks}

\begin{figure}[t]
	\centering
	\hspace{-1.2em}
	\includegraphics[width=1.05\linewidth]{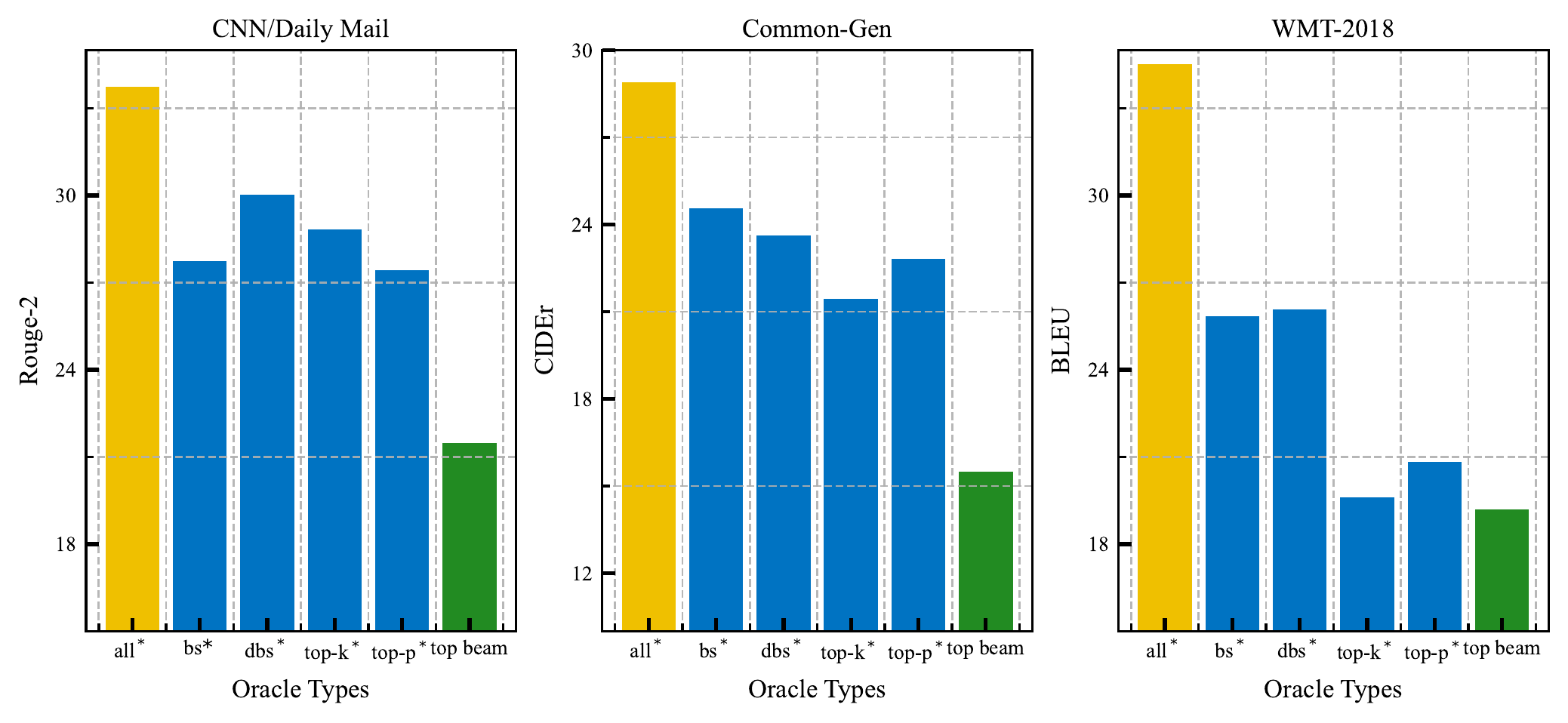}
	\caption{The comparisions between different decoders and oracle selections.}
	\label{fig:hist}
\end{figure}

To quantitatively understand how sub-optimal the default selections of decoding methods are, we present an empirical analysis in Fig.~\ref{fig:hist} .
Here we look at three typical NLG tasks: summarization (CNN/DM), machine translation (WMT18), and constrained text generation (CommonGen), with their popularly used base models: PEGASUS ~\cite{zhang2020pegasus}, Opus-MT~\cite{opus_mt}, and T5-large~\cite{t5}. 
We can see that the default selections (yellow bars in Fig.~\ref{fig:hist}; the top-beam generations) are much worse than the oracle selections from the top 15 candidate generations for each decoding method (blue bars).

Moreover, if we combine the results from the four methods as a larger candidate pool, then the performance (green bars) of these NLG models can be much improved. 
For example, the Rouge-2 score of PEAGUS can be improved by 57\% and the BLEU score for Opus-MT can be improved by nearly 80\%, compared to their top-beam performance. 
Simply put, the default selections (i.e., the generations with the highest decoding scores) are much worse than the best selections from a relatively small candidate pool. 
Therefore, we argue that it is of significant importance to rerank generation candidates in order to enhance the performance of LMs in NLG tasks.

Why do decoding algorithms often overlook generation candidates of better quality? The lower quality of default selections is often attributed to the exposure bias caused by the teacher-forcing paradigm in most auto-regressive models. 
Plus, the greediness of the search process and the randomness in sampling are also part of the reasons.
Re-ranking has been a simple yet effective post hoc approach to mitigate this gap. 
For instance, MLM-scoring~\cite{mlmscoring} uses an external LM such as BERT to estimate the quality of a candidate without any supervision.
SimCLS \cite{Liu2021SimCLSAS} trains a re-ranker using a simple contrastive training objective, which encodes the source text and each candidate output using the same encoder and scores each candidate based on the cosine similarity between the embeddings. 
Another successful approach is SummaReranker (SR)~\cite{Ravaut2022SummaRerankerAM}, which is trained to improve the re-ranker for multiple metrics simultaneously. 

\section{Additional Results on Conventional Tasks}
In this section, we evaluate the \ranker{} with conventional natural language generation (NLG) tasks.
Extensive experiments conducted on three NLG
tasks (i.e., summarization, translation, and constrained sentence generation) demonstrate that \ranker{} outperforms the baseline methods by a consistent margin and is also compatible with very large language models such as GPT-3 (text-davinci-003). 
\ranker{} not only  outperforms the previous state-of-the-art method SummaReranker on the summarization task, but also shows great generalization performance in the other two
NLG tasks, which are not evaluated previously. In
addition, our \ranker{} can be transferred
to improve GPT-3 results by 26.53\% and 11.65\%
for CommonGen and WMT18 (zh-en) respectively, even though our rerankers are not trained with any candidates decoded by GPT-3 models.

\subsection{Tasks and data creation}

We evaluate reranking methods on the following public dataset: CNN/DM, CommonGen, and WMT18 (zh-en). The data statistics of these benchmarks are in Table~\ref{tab:dataset_stat} (in Appendix).

\paragraph{CNN/DM} \cite{Hermann2015TeachingMT} is a dataset constructed from CNN and DailyMail websites. It is first used for machine-reading and comprehension, and later \citet{Nallapati2016AbstractiveTS} use it for abstractive summarization. Evaluation metrics are Rouge-1, Rouge-2, and Rouge-L.

\paragraph{CommonGen} \cite{lin2019commongen} is a dataset used for generative commonsense reasoning. It contains 79K commonsense descriptions where the language model is required to compose a realistically plausible sentence from given concepts. Evaluation metrics are BLEU and CIDEr. 

\paragraph{WMT2018} \cite{Bojar2018FindingsOT} is a well-known dataset for evaluate machine translation. Here we use the Chinese-English split for evaluation. Evaluation metrics are BLEU.

\subsection{Base models}

For the summarization task on CNN/DailyMail dataset, we use the famous PEGASUS-large~\cite{zhang2020pegasus} and BART-large~\cite{lewis2019bart}, which have exhibited great ability for abstractive summarization. We use the public fine-tuned checkpoint from Hugging face.
For the generative commonsense reasoning task on CommonGen dataset, we use the T5-large \cite{t5}. It's one of the vanilla baselines reported in \citet{lin2019commongen}. 
For the Chinese-English translation task on WMT18 dataset, we use the public pre-trained opus-mt checkpoint~\cite{Tiedemann2020OPUSMTB}.

\subsection{Evaluation setups}
\label{sec:evaluation_setup}

In this section, we talk about the training and testing paradigm of our reranker, including how we construct the training, validation, and testing dataset for our reranker, how we generate candidates, and what our experiment focuses on.
\begin{table}[t]

\begin{center}
    \scalebox{0.65}{
    \begin{tabular}{c c c c  c@{}}
        \toprule
         \textbf{Method} $\downarrow$ \textbf{Metric} $\rightarrow$ & R-1 & R-2 & R-L & Gain$_{R1}$   \\
        \midrule
        BART & 44.48 & 21.21 & 41.60 & - \\
        \textbf{PEGASUS} & 44.56 & 20.90 & 41.58 & - \\
        Gsum & 45.94 & 22.32 & 42.48 & -\\
        \midrule
        Gsum+RefSum & 46.18 & 22.36 & 42.91 & 1.18\%\\
        BART+SimCLS & 46.67 & 22.15 & 43.54 & 4.92\%\\
        PEGASUS+MLM-Scoring & 43.03 & 19.48 & 40.12 & -3.43\% \\
        PEGASUS+SummaReranker &47.16 & 22.55 & 43.87 & 5.83\% \\
        \hdashline
        PEGASUS+\textbf{PairReranker} (bubble) & 47.29 & 22.77 & 44.06 & 6.13\% \\
        PEGASUS+\textbf{PairReranker} (max wins) & 47.29 & 22.79 & 44.07 & 6.13\% \\
        PEGASUS+\textbf{PairReranker} (max logits) & \textbf{47.39} & \textbf{22.91} & \textbf{44.18} & \textbf{6.35}\% \\

        \bottomrule
        \toprule
        \textbf{GPT-3} (text-davinci-003) & 37.96 & 15.51 & 34.39 & - \\
        \textbf{GPT-3}-oracle   & 45.46 & 22.83 & 42.04 & 19.76\% \\
        GPT-3+MLM-Scoring & 38.13 & 15.09 & 34.32 & 0.45\% \\
        GPT-3+SummaReranker & 39.62 & 17.13 & 36.12 & 4.37\% \\
        \hdashline
        GPT-3+\textbf{PairReranker} (bubble) & 40.41 & 17.44 & 36.79 & 6.45\% \\ 
        GPT-3+\textbf{PairReranker} (max wins) & 40.37 & 17.46 & 36.76 & 6.35\% \\ 
        GPT-3+\textbf{PairReranker} (max logits) & \textbf{40.48} & \textbf{17.54} & \textbf{36.84} & \textbf{6.64}\% \\ 
        \bottomrule
    \end{tabular}}
    \caption{\label{tab:cnndm_results} Model performance on \textbf{CNN/DailyMail}.}
\end{center}
\end{table}

To construct the training dataset for the reranker, we need to ensure the base model used to generate candidates on the training dataset should never have seen these candidates. Otherwise, the reranker will be trained on the candidates with higher quality compared to the candidates that it will be tested on, which we found will result in fairly bad performance. Therefore, following \citet{Ravaut2022SummaRerankerAM}, we first fine-tune the original non-finetuned pre-trained model on half of the training dataset, which gives us 2 half-finetuned base models that each of them has only seen their own half of the training dataset. Then we use them to generate candidates on their un-seen half of the training dataset using the decoding method talked about before. These generated candidates together form a whole training dataset with generated candidates that resemble the quality during the inference stage. 

During the inference stage, we directly adopt the public checkpoints that have been finetuned on the whole training dataset. We generate candidates on the validation and test datasets with this public checkpoint, which constitutes the validation and testing datasets on which our reranker runs inference.
We use two decoding methods, beam search, and diverse beam search, in the experiments, following the prior work of SummaReranker. 
We generate 15 candidates for each decoding method for both training and inference.

\subsection{Main results}
\label{main results}


\paragraph{Overall performance in summarization.}

Following the training and testing paradigm stated in section \ref{sec:evaluation_setup}, we briefly report the test results on the CNN/DM dataset in Tab.~\ref{tab:cnndm_results}. With fine-tuned PEGASUS-large as the base model. our \textit{Max Logits} method improves the candidates' quality by 6.35\% in Rouge-1, which is higher than our baseline SummaReranker. Besides, the performance gains in other metrics like Rouge-2 (9.62\%) and Rouge-L (6.25\%) are also obviously better.

\paragraph{Can PairReranker generalize to other generation tasks?}
Yes. In order to test the task generalization ability of our method, we here report the test results on CommonGen and WMT2018 (zh-en) in Tab.~\ref{tab:commongen_results} and Tab.~\ref{tab:wmt2018_results}. From the data in the table, our method also improves the candidates' quality significantly after reranking. Our \textit{Max Logits} method obtain a 2.45\% performance gain in CIDEr on the CommonGen dataset and a 6.12\% performance gain in BLEU on the WMT2018 dataset. What's more, it's worth noting our \textit{bubble run} method achieves an even higher gain in CIDEr (2.91\%).

\begin{table}[t]
\vspace{-1em}
\begin{center}
    \scalebox{0.72}{
    \begin{tabular}{c c c c @{}}
        \toprule
          \textbf{Method} $\downarrow$ \textbf{Metric} $\rightarrow$  & BLEU & CIDEr  & Gain$_{\text{CIDEr}}$   \\
        \toprule
        T5-large & 14.62 & 15.48 &  - \\
        T5-large+MLM-Scoring & 14.04 & 14.12 & -8.79\% \\
        {T5-large+SimCLS} & 14.5 & 14.99 & -3.17 \\
        T5-large + SummaReranker & 14.13 & 15.29 & -1.23\% \\ 
        \hdashline
        T5-large + \textbf{PairReranker} (bubble)
        & 15.30 & \textbf{15.93} & \textbf{2.91\%} \\
        T5-large + \textbf{PairReranker} (max wins)
        & 15.29 & 15.91 & 2.78\% \\
        T5-large + \textbf{PairReranker} (max logits)
        & \textbf{15.40} & 15.86 & 2.45\% \\
    \bottomrule
        \toprule
        
        GPT-3 (text-davinci-003) & 11.85 & 11.12 & - \\
        GPT-3 + oracle  & 20.34 & 19.26 & 73.20\% \\
        GPT-3 + MLM-Scoring & 12.56 & 11.66 & 4.86\% \\
        GPT-3 + SummaReranker & 13.71 & 13.21 & 18.79\% \\
        \hdashline
        GPT-3 + \textbf{PairReranker} (bubble) & 14.39 & 13.85 & 24.55\% \\ 
        GPT-3 + \textbf{PairReranker} (max wins) & 14.32 & 13.76 & 23.74\% \\ 
        GPT-3 + \textbf{PairReranker} (max logits) & \textbf{14.63} & \textbf{14.07} & \textbf{26.53}\% \\ 
        \bottomrule
    \end{tabular}}
    \caption{\label{tab:commongen_results} Model performance on \textbf{CommonGen}. }
\end{center}
\end{table}

We also report the performance of SummaReranker on the two datasets. In contrast to the great performance on summarization, SummaReranker seems to fail to generalize well on other datasets. We also find that SummaReranker obtains a decreased gain on the CommonGen dataset (-1.23\% in CIDEr). The improvement on the translation dataset is not obvious (0.57\% in BLEU). We hypothesize that this is because of the average length of the candidates and the target text in these two datasets are all significantly smaller than the one in summarization (see in Tab.~\ref{tab:dataset_stat}). Therefore, the higher in-group similarity brought by the shorter length makes it harder for SummaReranker to capture their difference. On the contrary, our method with direct attention between a pair of candidates could easily tackle this problem.

\paragraph{Can PairReranker generalize other large-scale models like GPT-3?}
Yes. 
{After training on an expert dataset, our reranker could directly be applied to other models' outputs selection for the same task. To support this, we directly apply our three rerankers trained on the 3 main tasks respectively to the GPT-3 outputs with proper task-specific prompts.
} 
We report the performance gain in Tab.~\ref{tab:cnndm_results}, ~\ref{tab:commongen_results}, and~\ref{tab:wmt2018_results}. From the data reported in the table, we could see that the quality of the GPT-3 outputs is improved by a large margin compared to the average. Also, our performance gain is significantly larger than the baseline SummaReranker. For example, on the GPT-3 data points sampled from CNN/DM, our max logits method obtain a gain of 6.64\%, whereas SummaReranker only obtains a gain of 4.37\%. And on the CommonGen's, our method obtains a gain of 26.53\% and SummaReranker only obtains a gain of 18.79\%.

\begin{table}[!t]
\vspace{-1em}
\begin{center}
    \scalebox{0.72}{
    \begin{tabular}{c c c @{}}
        \toprule
          \textbf{Method} $\downarrow$ \textbf{Metric} $\rightarrow$  & BLEU & Gain     \\
        \midrule
        Opus-MT & 19.29 & - \\ 
        Opus-MT+MLM-Scoring & 16.35 & -15.24\% \\
        {Opus-MT+SimCLS} & 18.93 & -1.87\% \\
        Opus-MT+SummaReranker & 19.40 & 0.57\% \\
        \hdashline
        Opus-MT+\textbf{PairReranker (bubble)} & 20.36 & 5.54\% \\ 
        Opus-MT+PairReranker (max wins) & 20.30 & 5.24\% \\ 
        Opus-MT+\textbf{PairReranker (max logits)} & \textbf{20.47} & \textbf{6.12}\% \\ 

        \bottomrule
        \toprule
         GPT-3 (text-davinci-003) & 23.61 & - \\
        GPT-3 + oracle & 36.11 & 52.94\% \\
        GPT-3+MLM-Scoring & 23.98 & 1.57\% \\
        GPT-3+SummaReranker & 25.08 & 6.22\% \\
        \hdashline
        GPT-3+\textbf{PairReranker} (bubble) & 26.29 & 11.35\% \\ 
        GPT-3+\textbf{PairReranker} (max wins)& \textbf{26.36} & \textbf{11.65}\% \\ 
        GPT-3+\textbf{PairReranker} (max logits) & 26.19 & 10.93\% \\ 

        \hline
    \end{tabular}
    }
    \caption{\label{tab:wmt2018_results} Model performance on \textbf{WMT18 (zh-en)}.}
\end{center}
\end{table}

\begin{figure*}[!t]
    \includegraphics[width=1.0\linewidth]{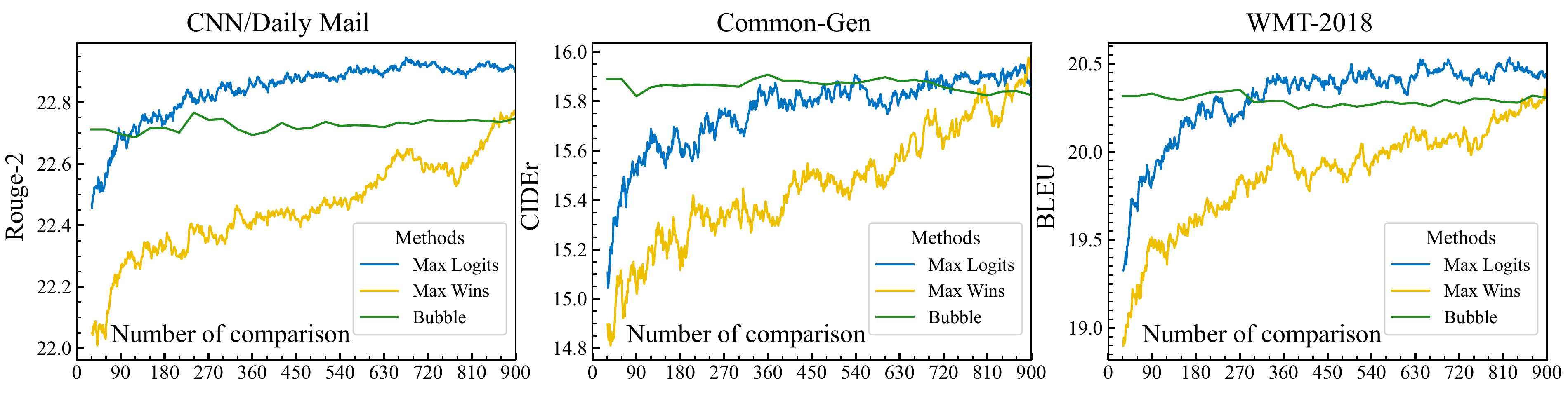}
    \caption{Efficiency trade-off with the number of pairwise comparisons}
    \label{fig:tradeoff_merged}
\end{figure*}

\paragraph{Can I make trade-offs between performance and number of comparisons?}


Yes. Due to the high cost of full comparison methods, it's necessary for us to study the trade-off between the model performance and the number of comparisons.
For full comparison methods, we first initialize matrix $M$ in Figure~\ref{fig:matrix} to be all zeroes, Then every time of the comparison, we fill a confidence cell that is zero before, then do the scoring and select the best one based on the current information in the matrix. For bubble run, we run multiple times of bubble run and select one that is chosen as the best one for the most times. Each bubble cost $N$ comparisons. The trade-off results are shown in Figure~\ref{fig:tradeoff_merged}.
From the results, we could see bubble run method could achieve high performance with little cost. However, as the number of comparisons increases, \textit{Max Logits} scoring methods will surpass the bubble run after a certain number of comparisons. 
{We contend that the bubble run method already reports a pretty good performance with $N-1$ times of comparison. Therefore, most of the time, bubble run is a more efficient way to apply. If you want to pursue the marginal improvements brought by more comparison, you can also apply \textit{Max Logits} method with parallel computing. 
}

\subsection{Model Further Study}
Due to the order of the input format, changing the position of candidate 1 and candidate 2 might also change the results (Sec. \ref{sec:architecture}). 
In practice, we found that by simply shuffling the order of candidate 1 and candidate 2, our reranker could be consistent with itself more than 90\% of the time. 

\begin{figure}[!h]
    \centering
    \includegraphics[scale=0.6]{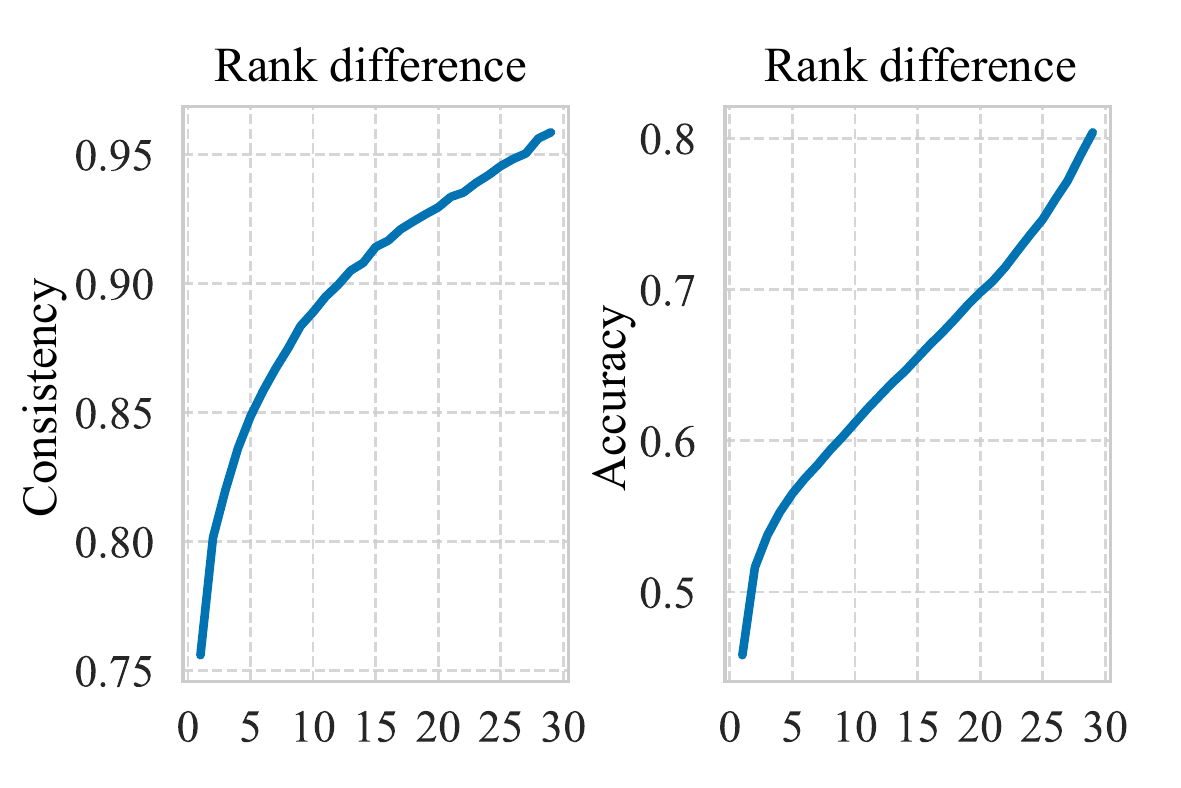}
    \caption{Consistency and accuracy analysis for CNN/Daily Mail Dataset}
    \label{cnndm_comparison_analysis}
\end{figure}

We analyze the model's relation between consistency as well as accuracy and the absolute pair rank difference. The results are presented in Figure~\ref{cnndm_comparison_analysis}. From the results, we could see that the model is better at classifying candidates with a higher absolute rank difference, because they are supposed to be more different.

\section{Dataset statistics}
We analyze the basic statistics, including the number of examples and the average words per example, of the 3 datasets. The data is presented in Table~\ref{tab:dataset_stat}
\begin{table}[!t]
\vspace{-2em}
\begin{center}
    \scalebox{0.75}{
    \begin{tabular}{c c c c   c c}
        \toprule
        \multirow{2}{3em}{Dataset} & \multicolumn{3}{c}{\# Examples} & \multicolumn{2}{c}{\# Words per example}\\
        ~ & Train & Val & Test & Source & Target \\
        \midrule
        CNN/DM & 287k & 13k & 11,490 & 856.56 & 70.05 \\
        CommonGen & 67k & 4k & 1,497 & 4.20 & 12.92 \\
        WMT18(zh-en) & 25m & 2k & 3,981 & 83.48 & 30.95 \\
        \bottomrule
    \end{tabular}}
    \caption{\label{tab:dataset_stat} Statistics of the three datasets.}
\end{center}
\end{table}

\onecolumn
\newpage
\section{ChatGPT Comparison Prompting Template (GPT-Rank)}
\begin{table}[htb]
    \centering
\scalebox{0.9}{
    \begin{tabular}{@{}c l@{}}
\hline
Template &  
\makecell[l]{
Instruction: \\
\$\{instruction\} \\
\\
Input: \\ 
\$\{input\} \\
\\
Candidate A: \\
\$\{candidate1\}\\
\\
Candidate B: \\
\$\{candidate2\}\\
\\
Given the instruction and input above, please compare the two candidates.\\
You only have 4 choices to output:\\
If you think A is better, please output: 1. A is better\\
If you think B is better, please output: 2. B is better\\
If you think both are good enough correctly give the answer, please output: 3. Same good\\
If you think both are bad and do not follow the instruction, please output: 4. Same bad\\
Do not output anything else except the 4 choices above.\\
Output your choice below:
}
\\
\hline
Comparison Option &
\makecell[l]{
1. A is better \\
2. B is better \\
3. Same good \\
4. Same bad
}\\
\hline

    \end{tabular}
}
    \caption{The template used for ChatGPT comparison ranking (GPT-Rank).}
    \label{tab:gpt_eval_template}
\end{table}

\end{document}